\documentclass[letterpaper, 10 pt, conference]{ieeeconf}  

\IEEEoverridecommandlockouts                              

\usepackage{booktabs} 
\usepackage{array}    
\usepackage{rotating} 
\usepackage{caption}  
\usepackage{graphicx} 
\usepackage{makecell} 
\usepackage{subcaption} 
\usepackage{multirow}
\usepackage{adjustbox}
\usepackage{amsmath}
\usepackage{amssymb}
\usepackage{siunitx}  
\usepackage{threeparttable} 

\title{\LARGE \bf
SupScene: Scene-Structured Overlap Supervision for Image Retrieval in Unconstrained SfM
}

\author{Xulei Shi$^{1}$, Maoyu Wang$^{1}$, Yuning Peng$^{1}$, Guanbo Wang$^{1}$, Xin Wang$^{1}$, Yifan Liao$^{1}$, Qi Chen$^{2}$, Pengjie Tao$^{1}$}

\begin{document}

\maketitle
\thispagestyle{empty}
\pagestyle{empty}


\begin{abstract}
Image retrieval is a critical step for reducing the quadratic cost of image matching in unconstrained Structure-from-Motion (SfM). Unlike generic image retrieval, however, the relevant goal of SfM is to identify geometrically matchable image pairs rather than merely semantically similar images. Prevailing methods are largely trained under anchor-centric tuple guidance, which organizes the training around isolated tuples and under-utilizes the dense, graded overlap structure naturally established within a SfM scene. In this work, we present SupScene, a scene-structured training framework that samples connected local subgraphs from SfM overlap graphs and jointly supervises all valid within-subgraph pairwise relations. To explicitly align the trained descriptor with geometric co-visibility, we further introduce an overlap-ordered objective that combines multi-similarity optimization with a continuous relative-overlap ranking term. In addition, the proposed framework is instantiated with a lightweight Structural Context Probe Pooling (SCPP) head that aggregates complementary structural responses into a compact global descriptor. Extensive experimental results on multiple benchmarks demonstrate that our method can significantly improve overall retrieval performance and enhance the completeness of downstream SfM reconstructions. Code and models are available at https://github.com/Suxilan/SupScene.
\end{abstract}
\section{INTRODUCTION}

Structure-from-Motion (SfM) is a fundamental task in the field of computer vision, photogrammetry and etc. Yet, scaling it to massive and unconstrained image collections remains challenging by the quadratic computational cost of exhaustive feature matching \cite{wang2019structure,Schonberger2016Colmap}. To mitigate this, image retrieval is often employed to pre-select candidate matchable pairs for downstream geometric verification \cite{nister2006bow}. However, unlike generic instance-based image retrieval, SfM aims to identify image pairs of geometric matchability rather than merely semantic relevance \cite{shen2018mirror,yan2021imvgcn,hou2023loip}. For instance, two images might depict the exact same architectural category or urban layout but observe disjoint regions. Conversely, two views with substantial geometric overlap may suffer from viewpoint, scale, or illumination while still forming a valid matchable pair. Therefore, this critical disparity makes SfM-oriented retrieval requires global descriptors that respond primarily to geometric co-visibility rather than visual resemblance alone.

\begin{figure}[t]
\centering
\includegraphics[width=\columnwidth]{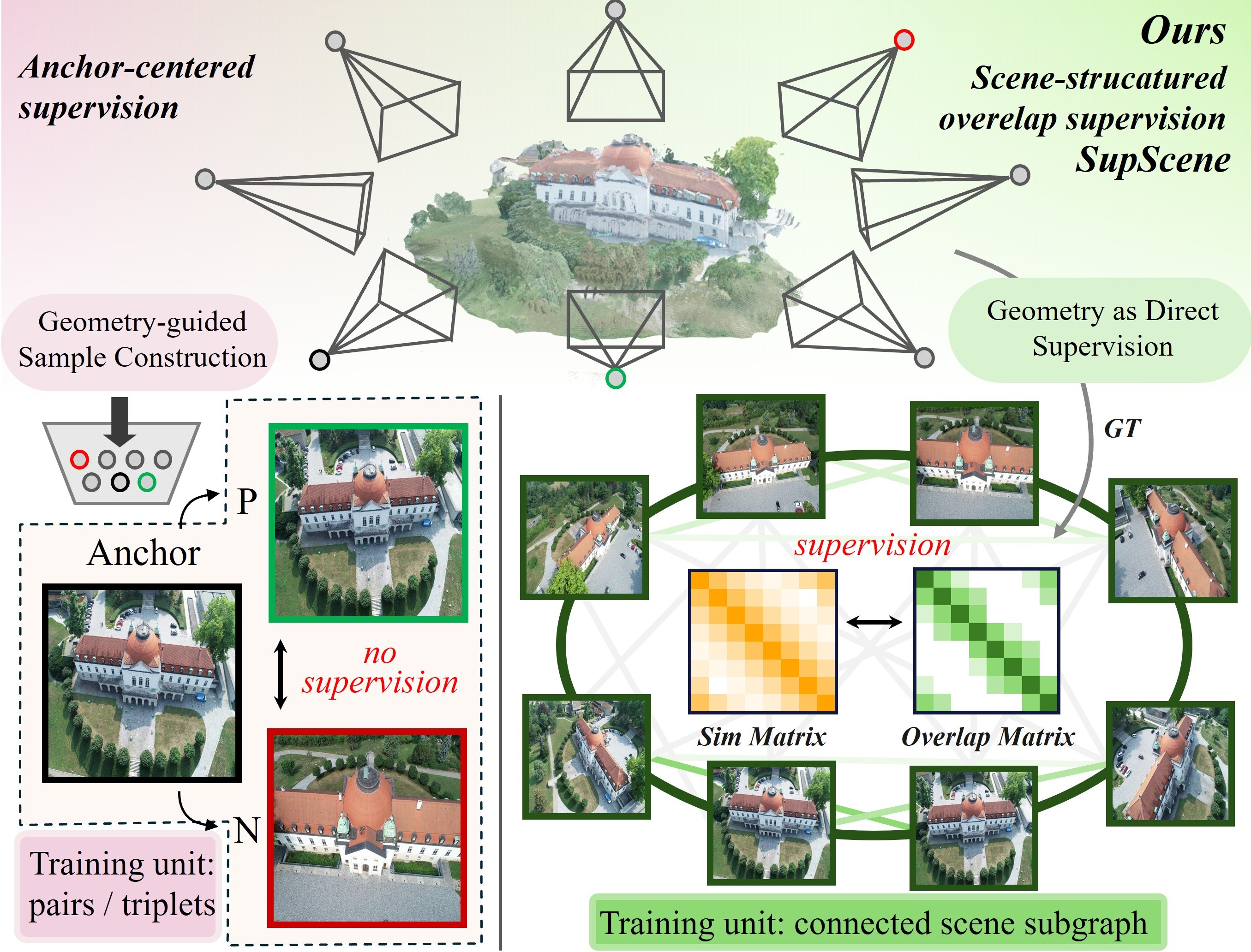}
\caption{Anchor-centric methods (left) vs. SupScene (right). }
\label{fig:overview}
\vspace{-20pt}   
\end{figure}

This critical distinction is also reflected in how supervision is organized during training. Early metric-learning approaches often used SfM pipelines mainly as an offline tool for mining hard training tuples in general instance retrieval \cite{gordo2016DIR,SiaMAC,radenovic2018GeM}. More recently, formulations tailored for matchable image retrieval have successfully integrated 3D reconstruction cues or photogrammetric data to construct more reliable training samples \cite{shen2018mirror,hou2023loip,liu2025uavpairs}. Despite their distinct objectives, both lines of work still organize supervision around anchor-centric training units, such as isolated pairs, triplets, or ranked lists. In these methods, geometry is mainly used to dictate which target images should be compared with a designated anchor, rather than to supervise scene relations directly. As illustrated in Fig.~\ref{fig:overview}(left), the loss is therefore defined by sampled anchor-centric comparisons, while the overlapping relations among the remaining images from the same scene are usually not modeled explicitly. In the context of SfM, however, a scene typically contains many images with measurable pairwise overlap. Once such highly structured scene relations are compressed into a small collection of discrete, anchor-centric samples, the supervision becomes fundamentally less informative about both the pairwise overlapping strength and the topological consistency of relations across multiple views.

To address this limitation, we introduce SupScene, an efficient scene-structured training framework that explicitly supervises within-scene relations, as illustrated in Fig.~\ref{fig:overview}(right). Instead of forming supervision from conventional anchor-centric samples, our SupScene organizes each training batch as a connected scene subgraph, so that valid within-subgraph relations can be jointly constrained in a single forward pass. On top of this training scheme, we design a joint objective that combines overlap-aware multi-similarity learning with a relative-overlap ranking term, encouraging descriptor similarity to preserve not only overlap-aware pair separation but also the relative order induced by overlap strength. We further adopt the framework with a lightweight Structural Context Probe Pooling (SCPP) head, which aggregates structural cues into a compact global descriptor while keeping the emphasis on the supervision scheme and optimization objective. In summary, our main contributions are as follows:

\begin{itemize}
\item We propose SupScene, a novel framework that reformulates SfM-oriented feature learning as scene-structured relation supervision over connected scene subgraphs.

\item  An overlap-aware joint loss is designed to unify a multi-similarity optimization with a relative margin ranking term. This objective explicitly penalizes ranking violations based on ground-truth overlap differences, ensuring the learned descriptor space strictly respects continuous geometric co-visibility.

\item We introduce Structural Context Probe Pooling (SCPP) as a structure-aware feature aggregator. By explicitly capturing structural responses from the backbone feature map, SCPP yields a compact global descriptor that achieves promising performance.
\end{itemize}

\section{RELATED WORKS}
This section reviews the evolution of image retrieval techniques, with a specific focus on their application within Structure-from-Motion (SfM) pipelines.
\subsection{SfM-Oriented Image Retrieval}

For over two decades, image retrieval has served as a pivotal front-end for scalable SfM, facilitating the efficient identification of candidate image pairs prior to computationally intensive local feature matching and two-view geometric verification. Early pipelines relied on handcrafted local features such as SIFT, SURF, and ORB \cite{lowe2004sift,bay2008surf,rublee2011orb}, combined with scalable indexing schemes of bag-of-visual-words and vocabulary trees \cite{csurka2004bow,philbin2007object,jiang2022bow}. To further improve distinctiveness and scalability, local descriptors were aggregated into compact global descriptors using VLAD and Fisher Vectors \cite{jegou2010VLAD,jegou2011aggregatingtpami,perronnin2010lFishvector}. This retrieval-based front-end design was often adopted by city-scale reconstruction and location-recognition systems, and remains a practical foundation in modern SfM pipelines \cite{Schonberger2016Colmap,schindler2007city,frahm2010building,agarwal2011building}. Together, these trials established image retrieval as an essential component for large-scale SfM rather than a peripheral acceleration heuristic.

\subsection{Deep Global Feature Aggregation for Image Retrieval}

The emergence of deep learning precipitated a transition from handcrafted encodings to learned global descriptors. Initial approaches utilized off-the-shelf CNN activations \cite{sharif2014cnn, babenko2015aggregating}, soon evolving into specialized aggregation mechanisms. Techniques like R-MAC, SPoC, and CroW \cite{tolias2016R-MAC, Babenko2015SPoC, kalantidis2016CroW} introduced spatial and channel-wise reweighting, while Generalized Mean (GeM) pooling \cite{radenovic2018GeM} and NetVLAD \cite{arandjelovic2016NetVLAD} formulated learnable pooling operators and differentiable clustering, respectively, becoming the de facto standards for global descriptor extraction.

Performance was further bolstered by metric-learning objectives and automated supervision. SfM-supervised fine-tuning \cite{radenovic2016sfm-retrieval}, automated labeling and hard-sample mining \cite{SiaMAC,radenovic2018GeM}, and architectures combining local and global cues \cite{cao2020eccv-gem,Yang2021dolg,Lee2022cvpr-gem,song2024ICRAglobalizing} have substantially improved descriptor discriminability. More recently, robust representations from vision foundation models (VFMs), particularly DINOv2 \cite{oquab2023dinov2}, have further enhanced the performance of image retrieval across diverse tasks \cite{keetha2023anyloc,Kordopatis-Zilos2025ILIAS}. However, while these foundational advances established deep global descriptors as the dominant standard for landmark retrieval, they also highlighted a critical distinction: retrieving valid pairs for SfM demands strict geometric overlap, differing fundamentally from generic instance retrieval which often relies on semantic resemblance.

\subsection{Geometry-Aware Supervision for Retrieval Learning}

To bridge the gap between semantic retrieval and geometric matchability, recent works inject 3D reconstruction cues into the training loop. Notable examples include MIRorR \cite{shen2018mirror}, which leverages reconstructed surfaces to refine triplet construction; IMvGCN \cite{yan2021imvgcn}, which introduces graph-based multi-view reasoning; and LOIP \cite{hou2023loip} and UAVPairs \cite{liu2025uavpairs}, which extend photogrammetry-driven supervision to complex aerial scenarios. Despite their sophisticated architectures and domain-specific priors, these methods still overwhelmingly cast the learning process into an anchor-centric supervision paradigm. By compressing rich geometric co-visibility into discrete, isolated tuples (e.g., pairs, triplets, or ranked lists \cite{gordo2016DIR, revaud2019aploss}), they inherently discard the continuous, hierarchical topological order among multiple overlapping views.

\begin{figure*}[t]
\vspace{-10pt}
\centering
\includegraphics[width=1.0\textwidth]{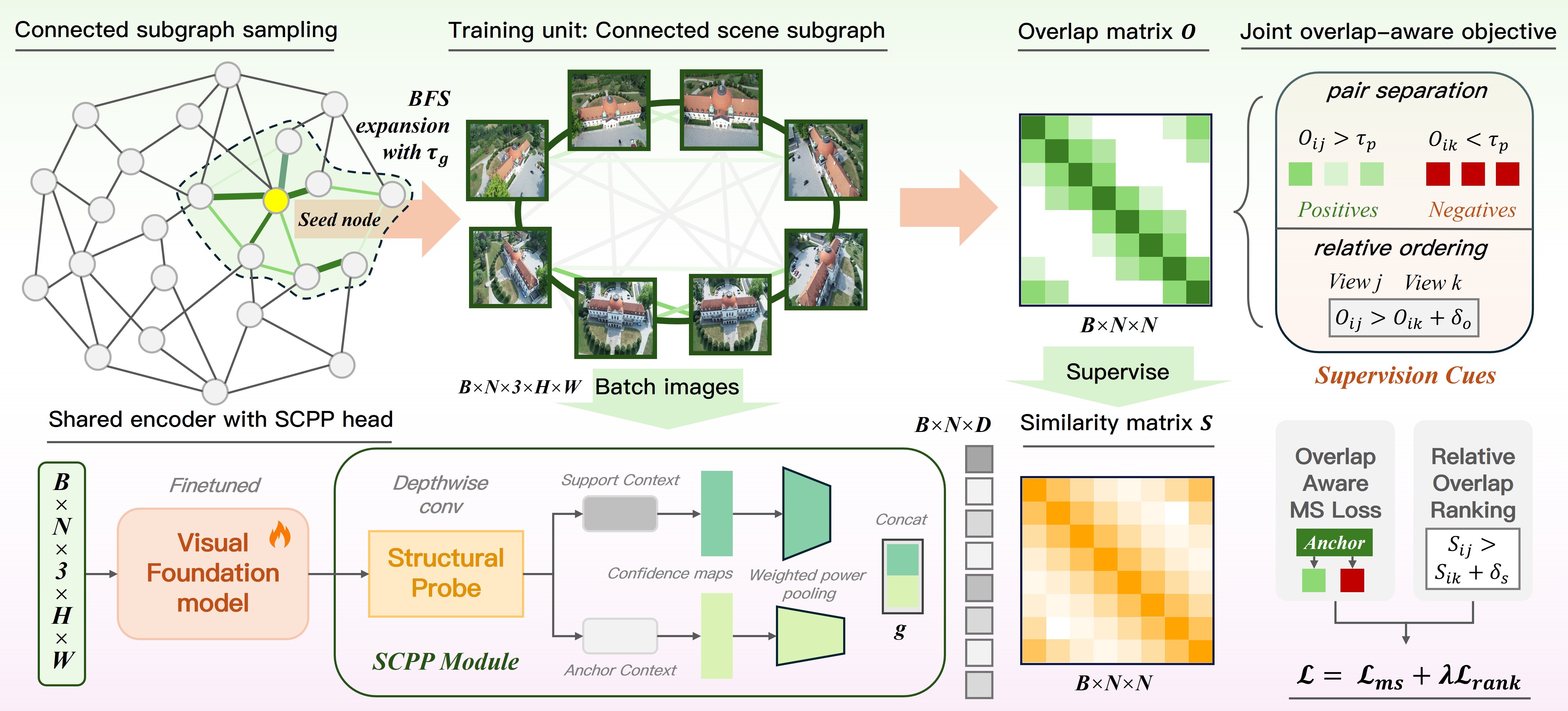}
\caption{Overview of the proposed training framework. A connected scene subgraph is first sampled from the overlap graph by thresholded BFS expansion, and represented by its induced overlap matrix $\mathbf{O}$. The sampled images are encoded by a shared visual backbone with the SCPP head to produce descriptors, whose similarity matrix $\mathbf{S}$ is optimized using a joint objective composed of overlap-aware pair separation and relative-overlap ranking.}
\label{fig:framework}
\vspace{-15pt}
\end{figure*} 

In contrast, our proposed SupScene framework fundamentally breaks away from this discrete sample-construction view. As outlined in the Introduction, it utilizes densely connected overlap subgraphs as holistic supervision targets. By jointly optimizing the entire similarity matrix against the explicit scene-level relational structure in a single forward pass, SupScene fine-tunes the backbone to strictly preserve continuous geometric topology, unlocking a more robust and highly efficient strategy for SfM-oriented retrieval.
\section{METHODOLOGY}
Given an unconstrained image collection, the goal of SfM-oriented retrieval is to reliably identify geometrically matchable pairs for downstream reconstruction. This objective dictates that the learned global descriptors possess the ability to explicitly capture the underlying 3D co-visibility. To this end, Sec. III-A introduces our scene-structured training framework, which organizes batches as connected local subgraphs and retains within-scene relations as the supervision target. Sec. III-B then details the Structural Context Probe Pooling (SCPP) head, a lightweight pooling layer designed to extract structural responses from strong visual backbones into a global descriptor. Finally, Sec.~III-C presents a joint overlap-aware objective that combines overlap-based pair separation with relative-overlap ranking.

\subsection{Scene-Structured Training over Overlap Graphs} 
\textbf{Overlap Graph Formulation.}
We model each 3D scene as an undirected, weighted overlap graph $\mathcal{G}=(\mathcal{V},\mathcal{E},w)$, where each node $v_i\in\mathcal{V}$ denotes an image and each edge $(v_i,v_j)\in\mathcal{E}$ represents a valid co-visibility relation. The edge weight $w_{ij}\in[0,1]$ carries the ground-truth overlap score (e.g., mesh-reprojection IoU) derived directly from the SfM pipeline \cite{shen2018mirror}. In practice, the sparse global overlap graph is symmetrized and self-loops are removed before subgraph extraction. For any sampled node set $\mathcal{V}_{\mathcal{B}}=\{v_{i_1},\ldots,v_{i_N}\}$ of size $N$, we construct its induced overlap matrix $\mathbf{O}\in[0,1]^{N\times N}$ as:  
\begin{equation}
\mathbf{O}_{jk}=
\begin{cases}
w_{i_j i_k}, & (v_{i_j},v_{i_k})\in\mathcal{E},\\
1, & j=k,\\
0, & \text{otherwise}.
\end{cases}
\end{equation}
This explicit matrix $\mathbf{O}$ serves as the geometric supervision target for the sampled subgraph. 

\textbf{Connected subgraph sampling.}
Instead of sampling isolated pairs or triplets as conventional methods do, SupScene utilizes a connected scene subgraph as the fundamental training unit. As depicted in the left panel of Fig.~\ref{fig:framework}, starting from a randomly selected seed image within a scene, we perform a breadth-first search (BFS) expansion along valid edges where the overlap weight exceeds a predefined propagation threshold $\tau_g$, and the expansion iteratively recruits connected nodes until a subgraph $\mathcal{G}_{\mathcal{B}}$ of size $N$ is formed. This thresholded propagation explicitly mines non-trivial samples directly from the scene's natural topology. It naturally surfaces informative within-scene samples of different difficulty levels, including strong positives, weak positives, and hard negatives, thereby providing rich relational cues for descriptor learning.

\textbf{Batch organization and forward pass.}
Given a batch of subgraphs, the input tensor is organized as $\mathbf{I}\in\mathbb{R}^{B\times N\times 3\times H\times W}$ and processed concurrently by a shared-weight visual foundation model. The spatial features are subsequently aggregated by our SCPP head (detailed in Sec.~III-B) to extract a set of global descriptors $\mathbf{X}\in\mathbb{R}^{B\times N\times D}$.  By computing the pairwise inner products between all descriptors within each subgraph, we obtain a predicted similarity matrix $\mathbf{S}\in[-1, 1]^{B\times N\times N}$. As illustrated in the right panel of Fig.~2, SupScene directly optimizes the entire similarity matrix $\mathbf{S}$ against the structural ground-truth matrix $\mathbf{O}$. To accommodate scenes with fewer images than $N$, a pair mask is applied during the loss computation to exclude padded or invalid relations.

\subsection{Structural Context Probe Pooling}
While modern Vision Foundation Models (VFMs) provide robust visual representations, their deep features are inherently biased toward high-level semantic abstractions. For SfM-oriented retrieval, the global descriptor must instead emphasize fine-grained structural cues that remain geometrically stable across drastic viewpoint changes. To achieve this without inflating the descriptor dimensionality(e.g., 1536-D), we design a lightweight \textbf{Structural Context Probe Pooling} (SCPP) head.

As illustrated in the bottom panel of Fig.~\ref{fig:framework}, given the backbone feature map $\mathbf{F}\in\mathbb{R}^{C\times H_f\times W_f}$, where $H_f$ and $W_f$ denote the spatial grid dimensions, SCPP first enhances local structural responses with a depthwise probe:
\begin{equation}
\mathbf{H}=\mathbf{F}+\mathrm{DWConv}(\mathbf{F}),
\end{equation}
where the residual form preserves the original representation while enhancing locally stable structures.

Based on the probed feature $\mathbf{H}$, two $1\times1$ convolutional projections generate complementary confidence maps,
\begin{equation}
\mathbf{C}_s=\sigma(\phi_s(\mathbf{H})), \qquad
\mathbf{C}_a=\sigma(\phi_a(\mathbf{H})),
\end{equation}
which respectively guide a \emph{support-context} branch and an \emph{anchor-context} branch. For each branch, we perform confidence-weighted power pooling:
\begin{equation}
\mathbf{v}(\mathbf{C},p)=
\left(
\frac{\sum_{u}\tilde{\mathbf{C}}(u)\,|\mathbf{F}(u)|^p}
{\sum_{u}\tilde{\mathbf{C}}(u)+\epsilon}
\right)^{\!1/p},
\end{equation}
where $\tilde{\mathbf{C}}$ denotes the confidence map normalized by its spatial mean, and $p$ controls the pooling sharpness. In our implementation, the two branches use different exponents, $p_s$ and $p_a$, to capture complementary structural responses.

The final descriptor is obtained by concatenating the $\ell_2$-normalized outputs of the two branches:
\begin{equation}
\mathbf{g}=
\big[
\mathrm{norm}(\mathbf{v}(\mathbf{C}_s,p_s)) \,;\,
\mathrm{norm}(\mathbf{v}(\mathbf{C}_a,p_a))
\big].
\end{equation}
This design keeps the aggregation head simple while allowing the descriptor to emphasize structurally informative regions under the proposed scene-structured supervision.

\subsection{Joint Overlap-Aware Objective}
To fully exploit the structural information encapsulated in the ground-truth matrix $\mathbf{O}$, our learning objective must align the predicted similarity space $\mathbf{S}$ with both the absolute existence and the continuous relative strength of geometric overlap. As depicted in Fig.~\ref{fig:framework}(right), our objective contains two complementary components: an overlap-aware pair separation term and a relative-overlap ranking term.

\textbf{Overlap-Aware Pair Separation.}
We first adapt the Multi-Similarity (MS) loss to establish a robust baseline for discriminating matchable views from unmatchable ones. For each reference image $i$, we first partition its within-subgraph pairs into positives and negatives according to an overlap threshold $\tau_p$:
\begin{equation}
\mathcal{P}(i)=\{j \mid O_{ij}\ge \tau_p\},\qquad
\mathcal{N}(i)=\{j \mid O_{ij}<\tau_p\}.
\end{equation}
Then the MS loss dynamically weights pairs based on their relative hardness, pulling positives together and pushing negatives apart:
\begin{equation}
\begin{aligned}
\mathcal{L}_{\mathrm{ms}} = \frac{1}{|\mathcal{V}|} \sum_{i\in\mathcal{V}} \Bigg[ 
&\frac{1}{\alpha} \log\!\Big(1+\sum_{j\in\mathcal{P}(i)} e^{\alpha(m-S_{ij})}\Big) \\
&+ \frac{1}{\beta} \log\!\Big(1+\sum_{k\in\mathcal{N}(i)} e^{\beta(S_{ik}-m)}\Big) \Bigg],
\end{aligned}
\end{equation}
where $\alpha$ and $\beta$ are scale factors, and $m$ is the base similarity margin. This term efficiently guarantees the fundamental separation between geometrically overlapping and disjoint pairs.

\textbf{Relative-Overlap Ranking.}
Binary pair separation alone does not preserve the relative order of overlap strength within a scene. We therefore introduce a ranking term over candidate triplets. For a reference image $i$ and two candidate views $j$ and $k$, if their ground-truth geometric overlap difference exceeds a predefined overlap margin $\delta_o$, we enforce their corresponding descriptor similarities to satisfy:
\begin{equation}
S_{ij}>S_{ik}+\delta_s ,
\end{equation}
where $\delta_s$ is the similarity margin. We penalize any violation of this geometric hierarchy using a margin ranking loss:
\begin{equation}
\mathcal{L}_{\mathrm{rank}}
=
\frac{1}{|\Omega|}
\sum_{(i,j,k)\in\Omega}
\max\!\big(0,\ \delta_s-S_{ij}+S_{ik}\big),
\end{equation}
with $\Omega = \{(i,j,k) \mid O_{ij} - O_{ik} > \delta_o\}$. This term explicitly preserves the continuous relative-overlap preference among candidate views from the same scene. The final objective jointly optimizes both constraints:
\begin{equation}
\mathcal{L}
=
\mathcal{L}_{\mathrm{ms}}
+
\lambda\,\mathcal{L}_{\mathrm{rank}},
\end{equation}
in our implementation, the ranking term is weighted by $\lambda$ and added as an auxiliary loss to the main MS loss.

\begin{table}[h]
\centering
\caption{Comparison of training strategy for capturing relations among $N$ images. Using triples as an example.}
\label{tab:comparison}
\resizebox{\columnwidth}{!}{%
\begin{tabular}{lcc}
\toprule
\textbf{Property} & \textbf{Anchor-Centric Tuples} & \textbf{SupScene (Ours)} \\
\midrule
Training Unit & Isolated $(a, p, n)$ & Connected Subgraph $\mathcal{G}_{\mathcal{B}}$ \\
Supervision Signal & Binarized (1 / 0) & $\mathbf{O}_{ij}$ + relative order \\
Hard Sample Mining & Explicit tuple & Sampled subgraph \\
\midrule
\multicolumn{3}{l}{\textit{Number of explicit tuples needed to fully model relations among $N=256$ images:}} \\
- Extreme Case\textsuperscript{*} & $65,024$ triplets & \textbf{1} dense matrix \\
- Balanced Case\textsuperscript{\dag} & $4,161,536$ triplets & \textbf{1} dense matrix \\
\midrule
\multicolumn{3}{l}{\textit{Naive image appearances at the encoder (uncached proxy):}} \\
- Extreme case\textsuperscript{*} & $195{,}072$ & \textbf{256} \\
- Balanced case\textsuperscript{\dag} & $12{,}484{,}608$ & \textbf{256} \\
\bottomrule
\multicolumn{3}{l}{\scriptsize \textsuperscript{*}Assume 1 positive and 254 negatives per anchor: $N \times 1 \times (N-2)$.} \\
\multicolumn{3}{l}{\scriptsize \textsuperscript{\dag}Assume 128 positives and 127 negatives per anchor: $N \times (N/2) \times (N/2 - 1)$.} \\
\end{tabular}%
}
\vspace{-0.2cm}
\end{table}

\textbf{Comparison to Anchor-Centric method.} 
To compare supervision granularity, consider a scene containing $N$ images. For an anchor-centric triplet formulation, let $P_i$ and $Q_i$ denote the numbers of positive and negative samples associated with anchor $i$, with $P_i+Q_i=N-1$. Explicitly exposing all anchor-centric triplet relations induced by this scene then requires
\begin{equation}
T=\sum_{i=1}^{N} P_i Q_i
\end{equation}
triplets, and the magnitude of $T$ depends on the per-anchor positive/negative distribution. As summarized in Table~\ref{tab:comparison}, when $N=256$, $T$ ranges from $65{,}024$ in the extreme case of one positive per anchor to $4{,}161{,}536$ when positives and negatives are approximately balanced. In contrast, SupScene encodes the $N$ images once as a connected scene subgraph and jointly optimizes the resulting overlap and similarity matrices in a single forward pass.

\section{EXPERIMENT}

\subsection{Experimental Settings}

\textbf{Datasets and evaluation metrics.}
We train and primarily evaluate our method on \textbf{GL3D} \cite{yao2020gl3d}, a large-scale benchmark for SfM and multi-view stereo. Following prior SfM-oriented retrieval works \cite{shen2018mirror,yan2021imvgcn}, image pairs with overlap larger than 0.25 are treated as effective matches for training and evaluation. The training split contains 110k images from 503 scenes, while the official GL3D test split of 40 scenes (8,914 images) is excluded from training. For retrieval, we report Recall@k and mean Average Precision (mAP)@k under full-database retrieval. We further evaluate downstream reconstruction on several large-scale scenes from \textbf{1DSfM} \cite{wilson_eccv2014_1dsfm}, a benchmark composed of crowd-sourced Flickr photo collections for large-scale unordered SfM, and report standard reconstruction statistics, including the number of registered images, mean track length, and mean reprojection error.

\textbf{Implementation Details.}
Our models are implemented in PyTorch and trained in distributed data-parallel mode on 8 NVIDIA RTX A6000 GPUs. Unless otherwise stated, we use \textbf{DINOv2-B} as the default visual backbone and fine-tune it with LoRA \cite{hu2022lora}, where both the rank and scaling factor are set to 16. For fair comparison with prior CNN-based retrieval methods, we additionally evaluate \textbf{ResNet-50} and \textbf{ResNet-101} under full fine-tuning. During training, images are resized to $224\times224$ and $322\times322$ for testing.
 
For scene-structured training, we set the BFS propagation threshold to $\tau_g=0.2$ to extract connected scene subgraphs of size $N=256$. In the joint objective, the positive pair threshold is defined as $\tau_p=0.25$, and the multi-similarity loss parameters are kept at their standard defaults($\alpha=2.0$, $\beta=50.0$, $m=0.5$). For the relative ranking branch, the margins $\delta_o=0.05$ and $\delta_s=0.05$ represent the minimum distinguishable units for geometric overlap and feature similarity, respectively. These values are derived from a statistical analysis of the training set, corresponding to the average minimum discriminability between the nearest and second-nearest neighbors. For the SCPP head, the contextual pooling exponents are empirically set to $p_s=1.3$ and $p_a=4.6$. The entire framework is optimized using AdamW with an initial learning rate of $1 \times 10^{-4}$, accompanied by a linear warm-up and cosine annealing decay. Models are trained with a batch size of 1 subgraph per GPU (yielding an effective batch size of 8 in distributed data-parallel mode) for a maximum of 10 epochs with early stopping.

\subsection{Comparison with Previous Works on GL3D}
We compare SupScene with representative global retrieval baselines on GL3D, including SiaMAC~\cite{SiaMAC}, NetVLAD~\cite{arandjelovic2016NetVLAD}, MIRorR~\cite{shen2018mirror}, and IMvGCN~\cite{yan2021imvgcn}. Because each query in SfM-oriented retrieval typically corresponds to many valid overlapping references, extremely small cutoffs are less informative; we therefore report Recall and mAP at $k\in\{25,100\}$. As shown in Table~\ref{tab:compare_gl3d}, the proposed framework performs strongly across both CNN and visual foundation backbones, and the CNN-based results already improve upon earlier geometry-aware retrieval methods.

Under the same backbone, we compare SCPP with GeM and NetVLAD within the same SupScene framework. SCPP remains consistently better than both alternatives, including under a PCA-compressed 768-D setting for direct comparison with GeM. The full 1536-D variant yields the best overall performance, suggesting that SCPP provides a stronger representation for SfM-oriented retrieval. Consequently, we deploy this top-performing model for downstream SfM reconstruction evaluations in the following subsection.

\begin{table}[ht]
\centering
\caption{\textbf{Comparison with previous retrieval methods on GL3D.} Baseline results are obtained using their official configurations. All SupScene variants are trained within our unified framework. \textbf{SCPP$^*$} denotes the PCA-compressed 768-D descriptor for fair dimensional comparison.}
\label{tab:compare_gl3d}
\renewcommand{\arraystretch}{1.2}
\setlength{\tabcolsep}{4pt} 
\small
\begin{tabular}{@{}>{\raggedright}p{1.7cm}>{\centering}p{1.6cm}>{\centering}p{0.8cm}cccc@{}}
\toprule[1.2pt]
\multirow{2}{*}{\textbf{Method}} & \multirow{2}{*}{\textbf{Backbone}} & \multirow{2}{*}{\textbf{Dim}} & \multicolumn{2}{c}{\textbf{Recall}} & \multicolumn{2}{c}{\textbf{mAP}} \\
\cmidrule(lr){4-5} \cmidrule(lr){6-7}
 & & & @25 & @100 & @25 & @100 \\
\midrule
SiaMAC\cite{SiaMAC} & CNN-Based & 512 & 59.4 & 88.6 & 61.9 & 66.2 \\
NetVLAD\cite{arandjelovic2016NetVLAD} & CNN-Based & 16384 & 58.8 &87.8 & 61.7 & 64.1 \\
MIRorR\cite{shen2018mirror} & CNN-Based & 256 & 61.1 & 90.3 & 64.2 & 73.4 \\
IMvGCN\cite{yan2021imvgcn} & CNN-Based & 2048 & 70.8 & 70.8 & — & — \\
\midrule
\multicolumn{7}{@{}l}{\textit{Ours (Trained within the SupScene Framework)}} \\
\midrule
GeM & Resnet50 & 1024 & 71.0 & 96.1 & 65.6 & 79.7\\
NetVLAD & Resnet50 & 32768 & 70.3 & 95.6 & 65.1 & 78.6\\
SCPP & Resnet50 & 2048 & 71.3 & 96.3 & 65.9 & 80.0 \\
SCPP & Resnet101 & 2048 & 72.2 & 96.7 & 67.4 & 80.3 \\
GeM & DINOv2 & 768 & 75.5 & 97.9 & 71.2 & 85.3 \\
NetVLAD & DINOv2 & 24576 & 75.4 & 97.7 & 70.9 & 84.8 \\
SCPP$^*$ & DINOv2 & 768 & \underline{78.2} & \underline{98.8} & \underline{72.7} & \underline{87.8} \\
SCPP & DINOv2 & 1536 & \textbf{78.6} & \textbf{99.0} & \textbf{73.5} & \textbf{88.5} \\
\bottomrule[1.2pt]
\end{tabular}
\vspace{-18pt}
\end{table}

\begin{table}[t]
\centering
\caption{\textbf{Downstream SfM reconstruction.} We report the number of registered images (Reg.), mean track length (Track), and mean reprojection error in pixels (Reproj.).}
\label{tab:sfm_result}
\renewcommand{\arraystretch}{1.08}
\setlength{\tabcolsep}{3.5pt}
\small
\resizebox{\columnwidth}{!}{%
\begin{tabular}{llcccc}
\toprule
\textbf{Scene} & \textbf{Method} & \textbf{\#Img} & \textbf{Reg.}$\uparrow$ & \textbf{Track}$\uparrow$ & \textbf{Reproj.}$\downarrow$ \\
\midrule
\multirow{5}{*}{Gendarmenmarkt}
& VocTree & \multirow{5}{*}{1463} & 970 & 6.45 & 0.705 \\
& SiaMAC  &                       & 977  & \textbf{6.48} & 0.722 \\
& MIRorR  &                       & 982  & 6.49          & 0.724 \\
& IMvGCN  &                       & \underline{997}  & 6.27          & \textbf{0.693} \\
& Ours    &                & \textbf{1081} & \underline{6.47} & \underline{0.704} \\
\midrule
\multirow{5}{*}{Madrid Metropolis}
& VocTree & \multirow{5}{*}{1344} & 407  & 6.94          & \textbf{0.609} \\
& SiaMAC  &                       & 467  & 7.08          & 0.617 \\
& MIRorR  &                       & \underline{478}  & 6.98 & 0.621 \\
& IMvGCN  &                       & 420  & \underline{7.16} & \underline{0.610} \\
& Ours    &                & \textbf{525} & \textbf{7.24} & 0.617 \\
\midrule
\multirow{5}{*}{Alamo}
& VocTree & \multirow{5}{*}{2915} & 880          & 11.72         & \textbf{0.630} \\
& SiaMAC  &                       & 904          & 11.95             & 0.661 \\
& MIRorR  &                       & \underline{925} & 11.89          & 0.658 \\
& IMvGCN  &                       & 862          & \textbf{12.02}    & \underline{0.632} \\
& Ours    &                & \textbf{984} & \underline{11.98} & 0.643 \\
\midrule
\multirow{5}{*}{Ellis Island}
& VocTree & \multirow{5}{*}{2587} & Fail          & Fail         & Fail \\
& SiaMAC  &                       & \underline{357}& 6.86       & \underline{0.808} \\
& MIRorR  &                       & Fail          & Fail         & Fail \\
& IMvGCN  &                       & 326          & \underline{6.96} & 0.811 \\
& Ours    &                & \textbf{431} & \textbf{7.13}       & \textbf{0.801} \\
\bottomrule
\end{tabular}%
}
\vspace{-20pt}
\end{table}

\begin{figure*}[t]
\centering
\includegraphics[width=1.0\textwidth]{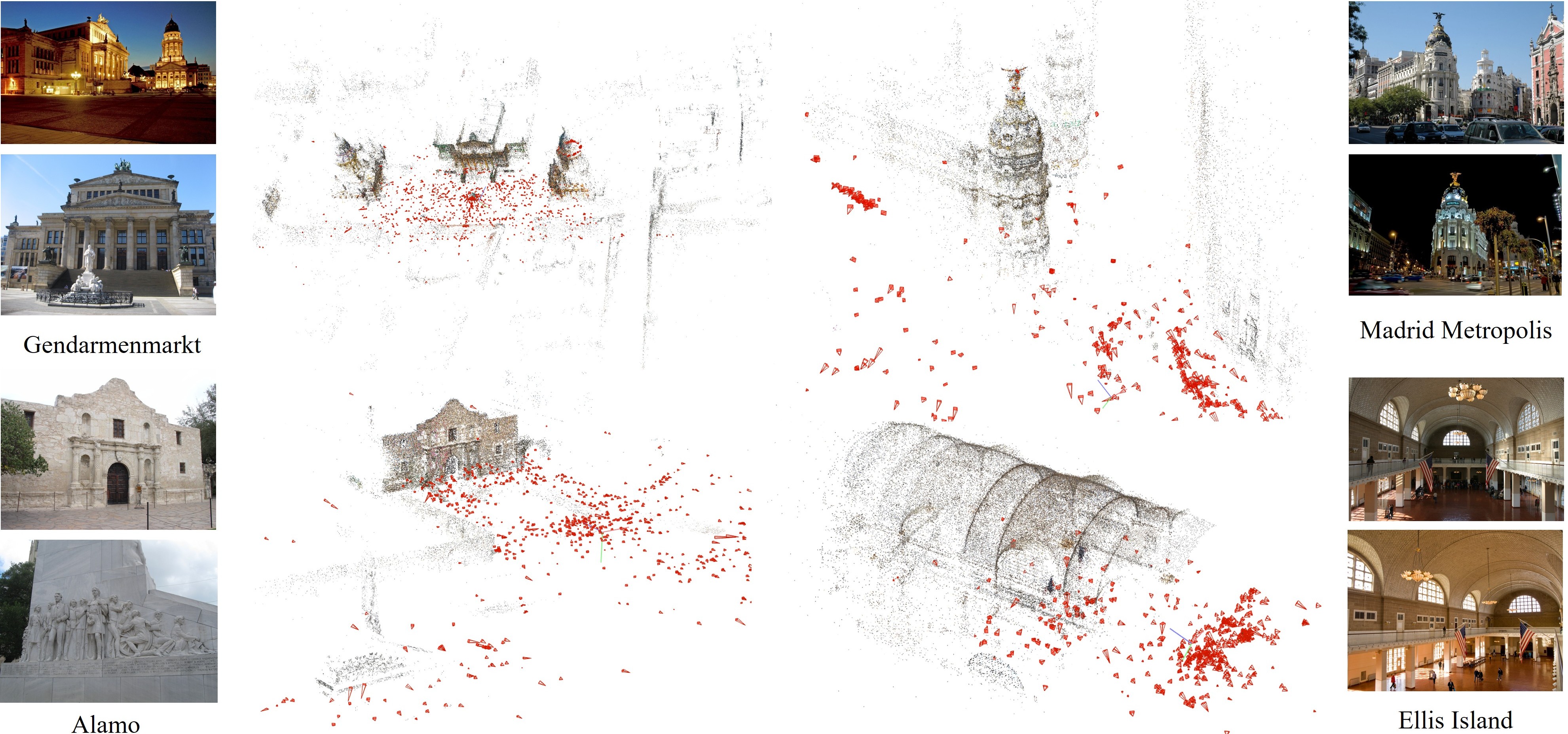}
\caption{\textbf{Qualitative SfM reconstruction results on representative 1DSfM scenes.}}
\label{fig:sfm_vis}
\vspace{-20pt}
\end{figure*} 

\begin{figure}[hb!]
\vspace{-15pt}
\centering
\includegraphics[width=\columnwidth]{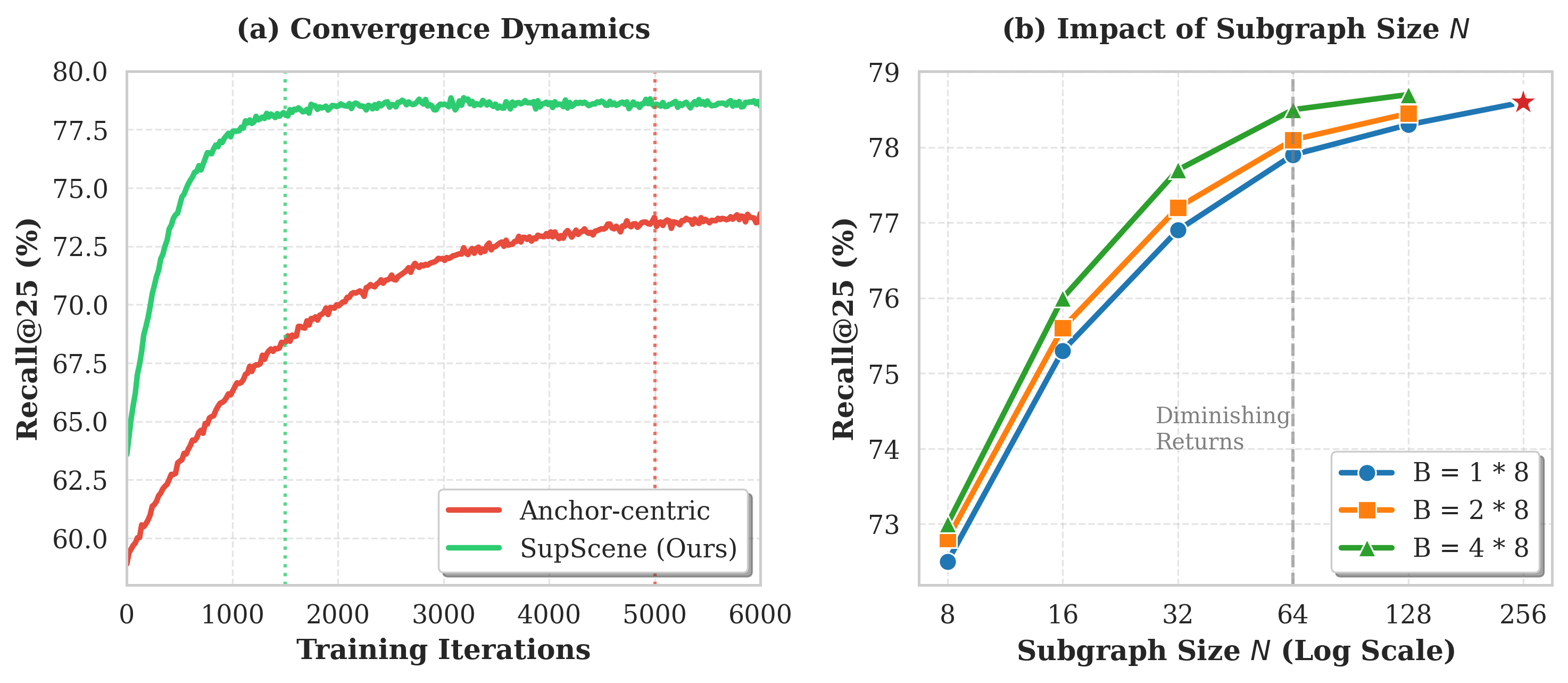}
\caption{\textbf{Training dynamics and configuration analysis.} 
\textbf{(a)} Convergence comparison on the GL3D dataset. 
\textbf{(b)} Impact of subgraph size $N$ and batch size $B$. $B$ denotes as $B \times 8$ to reflect the effective batch size across 8 GPUs in DDP mode. The red star ($\star$) denotes our adopted configuration.}
\label{fig:ablation_fig}
\vspace{-5pt}
\end{figure}

\subsection{Results on Downstream SfM Reconstruction}
We further evaluate the impact of our method on downstream 3D reconstruction using four challenging scenes from 1DSfM~\cite{wilson_eccv2014_1dsfm}. For each method, the top-100 retrieved candidates per query are used to construct the image-pair set for the same SfM pipeline. As reported in Table~\ref{tab:sfm_result}, SupScene consistently yields the largest number of registered images across all four scenes. While mean track length and reprojection error remain highly competitive, these two statistics are also influenced by the settings of   bundle adjustment. By contrast, the number of successfully registered images more directly reflects the quality of the retrieved candidate pairs for reconstruction. The consistent gains on diverse scene types, including landmarks, urban street environments, and indoor spaces, indicate that the proposed retrieval framework generalizes well across different geometric and visual conditions. This advantage is also visually reflected in Fig.~\ref{fig:sfm_vis}, where our method produces more complete reconstructed point clouds.  

\subsection{Ablation Studies}
To systematically validate the proposed training paradigm, objective, and aggregation head, we conduct ablation studies on GL3D. The quantitative results are summarized in Table~\ref{tab:ablation_core}, and the scene-structured training configurations are further analyzed in Fig.~\ref{fig:ablation_fig}.

\textbf{Effect of Scene-Structured Training Scheme.} 
We first compare the proposed SupScene training scheme against a strong anchor-centric baseline, denoted as \textbf{Anchor-c}, which follows the conventional Anchor-centered formulation with explicit $(a,p,n)$ tuple construction. Specifically, Anchor-c is optimized with multi-similarity loss under cross-batch hard mining, so that each anchor can interact with a large pool of positive and negative samples during training. This makes Anchor-c substantially stronger than a naive triplet baseline, and therefore provides a more meaningful reference for evaluating the proposed scene-structured supervision. To keep the comparison balanced in terms of sampled supervision and computational budget, we use a large batch size of 64 for Anchor-c, while SupScene processes one connected scene subgraph per GPU under the same model architecture setting. As shown in Block A of Table~\ref{tab:ablation_core}, replacing Anchor-centered tuple construction with scene-structured subgraph supervision consistently improves all retrieval metrics on both CNN and ViT backbones, indicating that the gain is not tied to a specific architecture.

To verify that the advantage of scene-structured supervision transfers beyond retrieval metrics, we further compare SupScene with the anchor-centric baseline on the representative 1DSfM scenes. As summarized in Table~\ref{tab:ablation_sfm}, SupScene consistently registers more images while maintaining comparable track length and reprojection error. This indicates that the gain from scene-structured supervision translates to better reconstruction coverage without sacrificing geometric consistency.

This advantage is also reflected in training dynamics. As shown in Fig.~\ref{fig:ablation_fig}(a), SupScene reaches near-saturated performance after only 1.5k iterations, whereas the anchor-centric baseline plateaus around 5k iterations. Fig.~\ref{fig:ablation_fig}(b) further illustrates that increasing the subgraph size $N$ yields larger gains than increasing the per-GPU batch size $B$, with improvements becoming marginal beyond $N=32$. Although jointly increasing both $N$ and $B$ can produce a slightly stronger Recall@25, the memory cost also grows substantially. In practice, adopting a single large subgraph per GPU provides the most favorable trade-off between retrieval accuracy and computational efficiency.

\begin{table}[t]
\centering
\caption{\textbf{Ablation studies on the GL3D-test dataset.} For the strong anchor-centric baseline, cross-batch hard mining is applied by default.}
\label{tab:ablation_core}
\renewcommand{\arraystretch}{1.15}
\setlength{\tabcolsep}{3pt} 
\small
\resizebox{\columnwidth}{!}{%
\begin{tabular}{@{}llccccc@{}}
\toprule[1.2pt]
\multirow{2}{*}{\textbf{Backbone}} & \multirow{2}{*}{\textbf{Head}} & \multirow{2}{*}{\textbf{Objective / Setting}} & \multicolumn{2}{c}{\textbf{Recall (\%)}} & \multicolumn{2}{c}{\textbf{mAP (\%)}} \\
\cmidrule(lr){4-5} \cmidrule(lr){6-7}
 & & & @25 & @100 & @25 & @100 \\
\midrule
\multicolumn{7}{@{}l}{\textit{A: Effect of Training Paradigm and Objective}} \\
\midrule
ResNet101 & SCPP & Anchor-c ($\mathcal{L}_{ms}$) & 68.6 & 94.5 & 63.8 & 75.1 \\
ResNet101 & SCPP & SupScene ($\mathcal{L}_{ms}$) & 70.8 & 96.2 & 66.1 & 78.8 \\
ResNet101 & SCPP & SupScene ($\mathcal{L}_{ms} \!+\! \mathcal{L}_{rank}$) & 72.2 & 96.7 & 67.4 & 80.3 \\
\cmidrule{1-7}
DINOv2 & SCPP & Anchor-c ($\mathcal{L}_{ms}$) & 74.1 & 97.4 & 69.5 & 82.6 \\
DINOv2 & SCPP & SupScene ($\mathcal{L}_{ms}$) & 76.3 & 98.2 & 71.8 & 86.5 \\
DINOv2 & SCPP & SupScene ($\mathcal{L}_{ms} \!+\! \mathcal{L}_{rank}$) & \textbf{78.6} & \textbf{99.0} & \textbf{73.5} & \textbf{88.5} \\
\midrule
\multicolumn{7}{@{}l}{\textit{B: Effect of SCPP Dual-Branch Design}} \\
\midrule
DINOv2 & GeM & Full & 75.5 & 97.9 & 71.2 & 85.3 \\
DINOv2 & SCPP (Supp.) & Full & 75.1 & 97.7 & 70.9 & 85.1 \\
DINOv2 & SCPP (Anch.) & Full & 76.3 & 98.2 & 71.8 & 86.6 \\
DINOv2 & SCPP (Dual) & Full & \textbf{78.6} & \textbf{99.0} & \textbf{73.5} & \textbf{88.5} \\
\bottomrule[1.2pt]
\end{tabular}%
}
\end{table}

\begin{table}[t]
\vspace{-5pt}
\centering
\caption{\textbf{Ablation studies on 1DSfM dataset.} }
\label{tab:ablation_sfm}
\renewcommand{\arraystretch}{1.1}
\setlength{\tabcolsep}{3pt}
\small
\resizebox{\columnwidth}{!}{%
\begin{tabular}{lccc|lccc}
\toprule
\multicolumn{4}{c|}{\textbf{Gendarmenmarkt}} & \multicolumn{4}{c}{\textbf{Madrid Metropolis}} \\
\midrule
Method & Reg.$\uparrow$ & Track$\uparrow$ & Reproj.$\downarrow$ & Method & Reg.$\uparrow$ & Track$\uparrow$ & Reproj.$\downarrow$ \\
\midrule
Anchor-c & 1024 & 6.41 & 0.709 & Anchor-c & 494 & 7.21 & \textbf{0.611} \\
SupScene & \textbf{1081} & \textbf{6.47} & \textbf{0.704} & SupScene & \textbf{525} & \textbf{7.24} & 0.617 \\
\midrule
\multicolumn{4}{c|}{\textbf{Alamo}} & \multicolumn{4}{c}{\textbf{Ellis Island}} \\
\midrule
Method & Reg.$\uparrow$ & Track$\uparrow$ & Reproj.$\downarrow$ & Method & Reg.$\uparrow$ & Track$\uparrow$ & Reproj.$\downarrow$ \\
\midrule
Anchor-c & 957 & 11.83 & 0.662 & Anchor-c & 389 & 7.09 & \textbf{0.799} \\
SupScene & \textbf{984} & \textbf{11.98} & \textbf{0.643} & SupScene & \textbf{431} & \textbf{7.13} & 0.801 \\
\bottomrule
\end{tabular}%
}
\vspace{-15pt}
\end{table}

\textbf{Effect of Relative-Overlap Ranking.} 
Table~\ref{tab:ablation_core} (Block A) shows that augmenting $\mathcal{L}_{ms}$ with the relative-overlap ranking term $\mathcal{L}_{rank}$ consistently improves both Recall and mAP across ResNet-101 and DINOv2-B. This suggests that incorporating relative-order optimization provides finer-grained supervision than threshold-based pair separation, encouraging the descriptors to capture subtle structural cues and align with continuous image overlap relations.

\textbf{Effect of SCPP Aggregation.} 
Table~\ref{tab:ablation_core} (Block B) evaluates the proposed SCPP head under the full SupScene objective. The dual-branch SCPP consistently outperforms GeM across all reported metrics, and ablating either branch causes a systematic drop in performance. These results indicate that the support-context and anchor-context branches capture complementary structural cues. Notably, the anchor-context branch alone is stronger than the support-context branch alone, while their combination achieves the best overall performance, validating the dual-branch design as a key component of the final descriptor head.

\subsection{Feature Visualization}
\begin{figure}[hb]
\vspace{-10pt}
\centering
\includegraphics[width=\columnwidth]{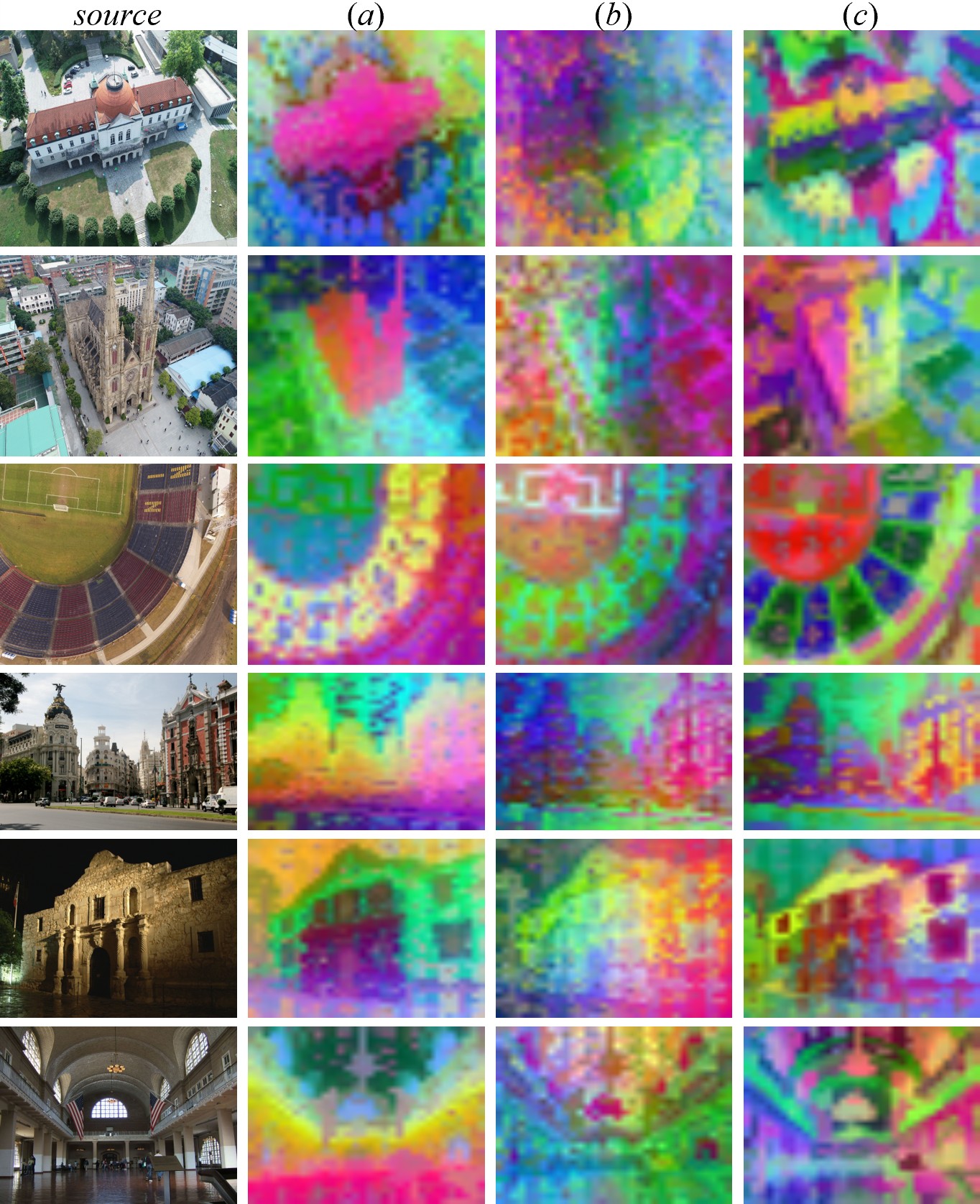}
\caption{\textbf{PCA visualization of spatial feature maps.} From left to right: input image, \textbf{(a)} original DINOv2 features, \textbf{(b)} features fine-tuned with the anchor-centric baseline, and \textbf{(c)} features fine-tuned with SupScene.}
\label{fig:feature_vis}
\vspace{-15pt}
\end{figure}

To gain deeper insight into the learned representations, we visualize the backbone's final patch-level feature maps via PCA projection into the RGB space. As shown in Fig.~\ref{fig:feature_vis}, SupScene consistently extracts highly structured and geometrically stable features across a diverse array of challenging scenarios, including aerial drone captures, extreme top-down perspectives, and night-time street scenes.

While original DINOv2 features primarily cluster into broad semantic regions, the anchor-centric baseline yields spatially diffuse responses. In contrast, SupScene yields more coherent feature organization and clearer responses around geometrically informative regions such as building contours, façades, and structural boundaries. This precise spatial delineation provides ideal inputs for our proposed SCPP head, allowing it to efficiently aggregate these prominent structural cues into a highly discriminative global descriptor. Ultimately, this qualitative evidence confirms that our scene-structured supervision successfully shifts features from generic semantic grouping toward structural representation.

\section{CONCLUSION}

This paper presents SupScene, a pioneering scene-structured training framework designed to advance image retrieval for unconstrained Structure-from-Motion (SfM). In contrast to conventional anchor-centric paradigms, SupScene explicitly optimizes dense topological relations within subgraphs. By synergizing multi-similarity learning with a continuous relative-overlap ranking objective, our approach compels the network to respect geometric co-visibility constraints, yielding robust and generalizable structural representations. Furthermore, we introduce the SCPP head to effectively aggregate multi-level structural responses into a compact global descriptor. Comprehensive evaluations confirm that SupScene not only achieves superior retrieval performance but also substantially boosts the registration completeness of downstream SfM pipelines.










\bibliographystyle{IEEEtran}
\bibliography{ref} 
\end{document}